\documentclass[twoside,11pt]{article} 
\usepackage{arxiv}

\usepackage{graphicx, epsfig, comment}
\usepackage{times,avant}

\usepackage{amssymb}
\usepackage{amsmath}

\usepackage{natbib}
\usepackage{subfigure}
\usepackage{algorithm}
\usepackage{algorithmic}
\usepackage[dvipsnames,usenames]{color}
\usepackage{multirow}

\def\RR{\mathbb{R}}

\def\cP{{\mathcal P}}
\def\cS{{\mathcal S}}

\def\SS{\mathbb{S}}
\def\RR{\mathbb{R}}

\DeclareMathOperator*{\argmin}{arg\,min}

\newcommand{\iid}{\stackrel{\mathrm{iid}}{\sim}}
\newcommand{\eqsvd}{\stackrel{\mathrm{svd}}{=}}

\newcommand{\mattwo}[4]
{
\begin{pmatrix} #1 & #2\\ #3 & #4\end{pmatrix}
}

\ShortHeadings{Adaptive Randomized Dimension Reduction on Massive Data}{Darnell, Georgiev, Mukherjee, \& Engelhardt} 
\firstpageno{1}

\begin{document}

\title{Adaptive Randomized Dimension Reduction on Massive Data}


\author{\name Gregory Darnell \email gdarnell@princeton.edu\\ 
\addr Lewis-Sigler Institute \\
Princeton University \\
Princeton, NJ 08544, USA
\AND 
\name Stoyan Georgiev \email sgeorg@stanford.edu\\ 
\addr Genetics Department \\
Stanford University \\
Palo Alto, CA 94305, USA
\AND 
\name Sayan Mukherjee \email sayan@stat.duke.edu\\ 
\addr Departments of Statistical Science\\
Mathematics, and Computer Science\\
Duke University\\
Durham, NC 27708, USA
\AND 
\name Barbara E Engelhardt \email bee@princeton.edu\\ 
\addr Department of Computer Science\\
Center for Statistics and Machine Learning\\
Princeton University \\
Princeton, NJ 08540, USA}


\maketitle

\begin{abstract} The scalability of statistical estimators is of increasing importance in modern applications. One approach to implementing scalable algorithms is to compress data into a low dimensional latent space using dimension reduction methods. In this paper we develop an approach for dimension reduction that exploits the assumption of low rank structure in high dimensional data to gain both computational and statistical advantages. We adapt recent randomized low-rank approximation algorithms to provide an efficient solution to principal component analysis (PCA), and we use this efficient solver to improve parameter estimation in large-scale linear mixed models (LMM) for association mapping in statistical and quantitative genomics. A key observation in this paper is that randomization serves a dual role, improving both computational and statistical performance by implicitly regularizing the covariance matrix estimate of the random effect in a LMM. These statistical and computational advantages are highlighted in our experiments on simulated data and large-scale genomic studies.
\end{abstract}


\section{Introduction}
In the current era of information, large amounts of complex high dimensional
data are routinely generated across science and engineering disciplines. Estimating and
exploring the underlying low-dimensional structure in data and using this latent structure to study scientific problems
is of fundamental importance in a variety of applications. As the size of the data sets
increases, the problem of statistical inference and computational feasibility become
inextricably linked. Dimension reduction is a natural approach to
summarizing massive data and has historically played a central role in data analysis,
visualization, and predictive modeling. It has had significant impact on both
statistical inference~\citep{Adcock1878,Edgeworth1884,Fisher1922,Hotelling1933,Young1941}, as well as on numerical analysis research and applications~\citep{Golub:1969,golub:van_loan,Gu_Eisenstat_1996,Golub00:gsvd};
for a recent review see~\citep{Mahoney:review}.
Historically, statisticians have focused on the study of theoretical properties
of estimators often in the context of asymptotically large number of samples. Numerical analysts and computational mathematicians,
on the other hand, have been instrumental in the development of useful and tractable
algorithms with provable stability and convergence guarantees.
Naturally, many of these algorithms have been successfully
applied to compute estimators grounded on solid statistical foundations.
A classic example of this interplay is principal component analysis (PCA)~\citep{Hotelling1933}. In PCA, an objective function is defined based on statistical
considerations about the sample variance in the data, which can then be efficiently
computed using a variety of singular value decomposition (SVD) algorithms
developed by the numerical analysis community. 

In this paper, we consider the problem of dimension reduction, focusing on the
integration of statistical considerations of \textit{estimation accuracy}
and out-of-sample prediction error of matrices with latent low-rank with computational considerations of run time
and \textit{numerical accuracy}.
The methodology that we develop builds on a classical approach to modeling large data,
which first compresses the data, minimizing the loss of relevant information, 
and then applies statistical estimators appropriate for small-scale problems.
In particular, we focus on dimension reduction via generalized eigendecomposition
as the means for data compression, and on out-of-sample residual error
as the measure of information loss.
The scope of this work includes applications to a large number of dimension reduction methods,
which can be implemented as solutions to truncated generalized eigendecomposition problems~\citep{Hotelling1933, LDA, Li:1991, lsir:2010}. In this paper our first focus is on
the increasing need to compute an SVD of massive data using randomized algorithms developed in
the numerical analysis community~\citep{Drineas:Ravi:Mahoney:2006,
Sarlos:random_projections:2006,Liberty:Woolfe,
Boutsidis:Drineas:2009, Rokhlin2008, Tropp2010} to  simultaneously reduce the dimension and regularize or control the impact of independent random noise.

The second focus in this paper is to provide efficient solvers for the linear mixed models that arise in statistical and quantitative genomics. In high-throughput genomics experiments, a vast amount of sequencing data is collected---on the order of tens of millions of genetic variants. The goal of genome-wide association studies is to test for a statistical association at each variant (polymorphic position) to a response of interest (e.g., gene expression levels or disease status) in a sample cohort. However, as the size of these genomic data and sample sizes continue to increase, there is an urgent need to improve the statistical and computational performance of standard tests.

It is typical to collect several thousand individuals for one study. These individuals may come from several genetically heterogeneous populations. It has been recognized since 2001~\citep{pritchard2001} that the ancestry makeup of the individuals in the study has great potential to influence study results---in particular, spurious associations arise when genetic variants with differential frequencies may appear to be associated with the biased response variable via latent population structure.

The earliest methods (e.g., genomic control) accounted for population structure by using covariate estimates to correct for these confounding signals. More recently, linear mixed models have been used successfully to correct spurious results in the presence of population structure. LMMs have been shown to improve power in association studies while reducing false positives~\citep{yang2014}. However, mixed models incur a high computational cost when performing association studies because of the computational burden of computing a covariance matrix for the random effect controlling for population structure. Significant work has gone into mitigating such costs using spectral decompositions for efficient covariance estimation~\citep{kang2008,kang2010,yang2011,zhou2012,listgarten2012}.

In this work, we show, using simulations of genomic data with latent population structure and real data from large-scale genomic studies that adaptive randomized SVD (ARSVD) developed here performs with both computational efficiency and numerical accuracy. Under certain settings, we find that the LMM using ARSVD outperforms current state-of-the-art approaches by implicitly performing regularization on the covariance matrix.

There are three key contributions of this paper:
\begin{enumerate}
\item[(i)] We develop an adaptive algorithm for randomized
singular value decomposition (SVD) in which both the number of
relevant singular vectors and the number of
iterations of the algorithm are inferred from the data based on informative statistical criteria.
\item[(ii)] We use our adaptive randomized SVD (ARSVD) algorithm to construct truncated generalized
eigendecomposition estimators for PCA and linear mixed models (LMMs)~\citep{listgarten2012,zhou2012}.
\item[(iii)] We demonstrate on simulated and real data examples that the randomized
estimators provide a computationally efficient solution, and, furthermore, 
often improve statistical accuracy of the predictions. We show that in an over-parametrized situation this improvement in accuracy is due to implicit regularization imposed by the randomized approximation.
\end{enumerate}
In Section~\ref{sec:methods}, we describe the adaptive randomized
SVD procedure we use for the various dimension reduction methods.
In Section~\ref{sec:gegin}, we provide randomized estimators for
linear mixed models used in quantitative genetics. 
In Section~\ref{sec:results}, we validate the proposed
methodology on simulated and real data and compare our approach
with state-of-the-art approaches. In particular, we show results from our approach for estimating low dimensional geographic structure in genomic data and for genetic association mapping applications.

\section{Randomized algorithms for dimension reduction}
\label{sec:methods}
In this section 
we develop algorithmic extensions for PCA.
We state an algorithm that
provides a numerically efficient and statistically robust estimate of the highest
variance directions in the data using a randomized algorithm for singular
value decomposition (Randomized SVD)~\citep{Rokhlin2008, Tropp2010}.
In this problem, the objective is linear unsupervised dimension reduction with
the low-dimensional subspace estimated via an eigendecomposition.
Randomized SVD will serve as the core computational engine for the other estimators we develop in this paper. 

\subsection{Notation}
Given positive integers $p$ and $d$ with $p \gg d$, $\RR^{p \times d}$ denotes the class of all matrices with real entries of dimension $p \times d$,
and $\SS^{p}_{++}$ ($\SS^{p}_{+}$) denotes the sub-class of
symmetric positive definite (semi-definite)
$p \times p $ matrices. For $B \in \RR^{p \times d}$, span($B$) denotes the
subspace of $\RR^p$ spanned by the columns of $B$. A \textit{basis matrix}
for a subspace $\cS$ is any full column rank matrix $B \in \RR^{p \times d}$
such that $\cS=\text{span}(B)$, where $d=\text{dim}(\cS)$. In the case of sample
data $X$, eigen-basis($X$) denotes the orthonormal left eigenvector basis.
Denote the data matrix by $X=(x_1,\ldots,x_n)^T \in \RR^{n \times p}$, where each
sample $x_i$ is a assumed to be generated by a $p$-dimensional probability
distribution $\cP_{_X}$. In the case of supervised dimensions reduction, denote the
response vector to be $Y \in \RR^{n}$ (quantitative response) or $Y \in \{1,\ldots,C\}$
(categorical response with $C$ categories), and $Y \sim \cP_{_Y}$. The data and
the response are assumed to have a joint distribution $(X, Y) \sim \cP_{_{X \times Y}}$.

\subsection{Computational considerations}
\label{sec2:computational}
The main computational tool we use is a randomized algorithm for approximate eigendecompositon, which
factorizes a $n \times p$ matrix of rank $r$ in time
${\cal O}(npr)$ rather than the ${\cal O}(np \times \mbox{min}(n,p))$ time required
by approaches that do not take advantage of the special structure
in the input. This is particularly relevant to statistical applications
in which the data is high dimensional but reflects a highly constrained process,
e.g., from biology or finance applications, which suggests it has low
intrinsic dimensionality, i.e., $r \ll n < p$. 
An appealing characteristic of our randomized algorithm
is the explicit control of the trade-off between estimation accuracy
relative to the exact estimates and computational efficiency.
Rapid convergence to the exact estimates was shown both empirically as well as in theory in
\citep{Rokhlin2008}.

\subsection{Statistical considerations}
\label{sec2:stats}
A central concept in this paper is that the randomized approximation
algorithms that we use for statistical inference implicitly impose regularization
constraints.
This idea is best understood by putting the randomized
algorithm in the context of a statistical model.
The numerical analysis and theoretical computer science perspective is deterministic and focuses on optimizing the discrepancy between the
randomized and the exact solution, which is estimated and evaluated
on a fixed data set.
However, in many practical applications, the data are noisy random
samples from a population. Hence, when the
interest is in dimension reduction, the relevant error comparison
is between the true subspace that captures information about
the population and the corresponding algorithmic estimates. The
true subspace is typically unknown, so we can validate the estimated
subspace by quantifying the out-of-sample generalized performance.
A key parameter of the randomized estimators, described in detail
in Section~\ref{sec:rsvd}, is the number of power iterations used to
estimate the span of the data. Increasing values for this parameter
provide intermediate solutions to the factorization problem, which
converges the to the exact answer~\citep{Rokhlin2008}. Fewer iterations
correspond to reduced run time but also to larger deviation from
the sample estimates and hence stronger regularization.
We show in Section \ref{sec:results} the regularization effect of
randomization on both simulated and real data sets. We argue that, in certain contexts, fewer iterations may be justified both by computational and statistical considerations.

\subsection{Adaptive randomized low-rank approximation}\label{sec:rsvd}
In this section we provide a brief description of a randomized estimator for the best low-rank matrix
approximation, introduced by \citep{Rokhlin2008, Tropp2010}, which combines random projections
with numerically stable matrix factorization. We consider this
numerical framework as implementing a computationally efficient
shrinkage estimator for the subspace capturing the largest variance
directions in the data, particularly appropriate when applied to input
matrix of (approximately) low rank. Detailed discussion of the estimation accuracy of Randomized SVD
in the absence of noise is provided in \citep{Rokhlin2008}.
The idea of random projection was first developed as a proof
technique to study the distortion induced by low dimensional
embedding of high-dimensional vectors in the work of
\citep{Johnson1984} with much literature simplifying and sharpening
the results \citep{JLS:Frankl:1987, Indyk:1998,JLS:simplified_proof:1999, Achlioptas:2001}.
More recently, the theoretical computer science and the numerical analysis
communities discovered that random projections can be used for
efficient approximation algorithms for a variety of applications
\citep{Drineas:Ravi:Mahoney:2006, Sarlos:random_projections:2006, Liberty:Woolfe,
Boutsidis:Drineas:2009, Rokhlin2008, Tropp2010}.
We focus on one such approach proposed by \citep{Rokhlin2008, Tropp2010} which
targets the accurate low-rank approximation of a given large data matrix $X \in \RR^{n \times p}$.
In particular, we extend the randomization methodology to the noise setting, in which the
estimation error is due to both the approximation of the low-rank structure in $X$,
as well as the added noise. A simple working model capturing this scenario is
as follows: $X = X_{d^{*}} + E$, where $X_{d^{*}} \in \RR^{n \times p}$, rank $(X_{d^{*}})=d^{*}$
captures the low dimensional signal, while $E$ is independent additive noise.

\paragraph{Algorithm.}
Given an upper bound on the target rank $d_{\max}$ and on the number of necessary
power iterations $t_{\max}$ ($t_{\max} \in \{5,\ldots 10\}$ is sufficient in most cases),
the algorithm proceeds in two stages: (1) estimate a basis for the range
of $X_{d^{*}}$, (2) project the data onto this basis and apply SVD: \\ \vspace{.1in}

\centerline{Algorithm: \textit{Adaptive Randomized SVD}(X, $t_{\max}$, $d_{\max}$, $\Delta$) }\label{alg:arsvd}
\begin{enumerate}
\item[(1)] Find orthonormal basis for the range of $X$;
\begin{enumerate}
\item[(i)] Set the number working directions: $\ell = d_{\max} + \Delta$;
\item[(ii)] Generate random matrix: $\Omega \in \RR^{n \times \ell}$ with $\Omega_{ij} \stackrel{iid}{\sim} \mbox{N}(0,1)$;
\item[(iii)] Construct blocks: $F^{(t)} = X X^T F^{(t-1)}$ with $F^{(0)} = \Omega$ for $t \in \{1,\ldots,t_{\max}\}$;
\item[(iv)] Select optimal block $t^{*} \in \{1,\ldots,t_{\max}\}$ and rank estimate $d^{*} \in \{1,\ldots, d_{\max}\}$,
using a stability criterion and Bi-Cross-Validation (see Section \ref{rankest_unsup});
\item[(v)] Estimate basis for selected block using SVD: $\ F^{(t^{*})} = Q R \in \RR^{n \times l},\ \ \ \ \ Q^{T}Q=I$;
\end{enumerate}
\item[(2)] Project data onto the range basis and compute SVD;
\begin{enumerate}
\item[(i)] Project onto the basis: $B = X^{T} Q \in \RR^{p \times \ell} $;
\item[(ii)] Factorize: $B \eqsvd U \Sigma W^{T}$, where $\Sigma = \text{diag(}\sigma_1,\ldots ,\sigma_{\ell}\text{)}$;
\item[(iii)] Rank $d^{*}$ approximation: $\widehat{X}_{d^{*}} = U_{d^{*}} \Sigma_{d^{*}} V_{d^{*}}^{T}$\\
$U_{d^{*}}= (U_1 | \ldots | U_{d^{*}} ) \in \RR^{n \times d^{*}} $\\
$\Sigma_{d^{*}}= \text{diag(}\sigma_1,\ldots ,\sigma_{d^{*}}\text{)} \in \RR^{d^{*} \times d^{*}} $\\
$V_{d^{*}} = Q \times (W_1|\ldots | W_{d^{*}}) \in \RR^{p \times d^{*}} $;
\end{enumerate}
\end{enumerate}

In stage (1) we set the number of working directions $\ell=d_{\max} + \Delta$ to be the sum of the upper bound
on the rank of the data, $d_{\max}$ and a small oversampling parameter $\Delta$, which ensures more stable
approximation of the top $d_{\max}$ sample variance directions; the estimator tends to be robust to changes in
$\Delta$, so we use $\Delta=10$ as a suggested default.
In step (1.iii) the random projection matrix $\Omega$ is applied to powers of $X X^T$ to
randomly sample linear combinations of eigenvectors of the data weighted by powers
of the eigenvalues:
\begin{equation*}
\underbrace{F^{(t)}}_{n \times \ell} = (XX^T)^{t}\Omega = US^{2t}U^T\Omega = US^{2t} \Omega^*, \quad \mbox{ where } X \eqsvd U S V^T.
\end{equation*}
The purpose of the power iterations is to increase the decay of
the noise portion of the eigen-spectrum while leaving the eigenvectors
unchanged. This is effectively a form of regularization as we are shrinking eigenvalues.
Note that each column $j$ of $F^{(t)}$ corresponds
to a draw from a $n$-dimensional Gaussian distribution:
$F^{(t)}_j \sim N(0, U S^{4t} U^T)$, with the covariance structure
more strongly biased towards higher directions of variation as
$t$ increases. This type of regularization is analogous to the
local shrinkage term developed in prior work~\citep{ScottPolson}. In step (iv),
we select an optimal block $F^{(t^{*})}$ for $t^{*} \in \{1,\ldots,t_{\max}\}$
and estimate an orthonormal basis for the column space.
In previous work~\citep{Rokhlin2008}, the authors assumed fixed target rank $d^{*}$
and approximated $X$, rather than $X_{d^{*}}$.
They showed that the optimal strategy for this purpose is to set $t^{*}=t_{\max}$, which typically
achieves excellent $d^{*}$-rank approximation accuracy for $X$,
even for relatively small values of $t_{\max}$.
In this work we focus on the noisy case, where $E\ne 0$, and propose
to adaptively set both $d^{*}$ and $t^{*}$, aiming to optimize
the generalization performance of the randomized estimator.
The estimation strategy for $d^{*}$ and $t^{*}$ is described in detail
in Section~\ref{rankest_unsup}.

In stage (2) we rotate the orthogonal
basis $Q$ computed in stage (1) to the canonical eigenvector
basis and scale according to the corresponding eigenvalues.
In step (2.i) the data is projected onto the low dimensional
orthogonal basis $Q$. Step (2.ii) computes the exact SVD in the
projected space.

\paragraph{Computational complexity.}
The computational complexity of the randomization step is 
${\cal O}(np \times d_{\max} \times t_{\max})$ and the factorizations in the 
lower dimensional space have complexity
${\cal O}(n p \times d_{\max} + n \times d^2_{\max} )$.
With $d_{\max}$ small relative to $n$ and $p$,
the run time in both steps is dominated by the multiplication by
the data matrix; in the case of sparse data, fast multiplication
can further reduce the run time. We use a normalized version
of the above algorithm that has the same run time complexity
but is numerically more stable~\citep{normalized_rsvd:2010}.

\subsubsection{Adaptive method to estimate $d^{*}$ and $t^{*}$}\label{rankest_unsup}
We propose to use ideas of stability 
under random projections in combination with cross-validation
to estimate the intrinsic dimensionality of the dimension reduction
subspace $d^{*}$ as well as the optimal value of the eigenvalue 
shrinkage parameter $t^{*}$.

\paragraph{Stability-based estimation of $d^{*}$:}
First, we assume the regularization parameter $t$ is fixed and describe 
the estimation of the rank parameter $d^{*}(t)$, using a stability
criterion based on random projections of the data.
We start with rough upper-bound estimate $d_{\max}$ for the dimension 
parameter $d^{*}$. We then apply a small number ($B = 5$) of independent Gaussian random projections $\Omega^{(b)} \in \RR^{n \times d_{\max}}$, $\Omega^{(b)}_{ij} \stackrel{iid}{\sim} \mbox{N}(0,1)$, for $b \in \{1,\ldots,B\}$. Given the projections we compute an estimate of the 
eigenvector basis of the column space onto the projected data. We then denoise the estimate by raising all the eigenvalues to the power $t$:
\begin{equation*}
U_{b}^{(t)} \equiv (U_{b1}^{(t)} | \ldots | U_{bd}^{(t)} ) = \mbox{SVD}[(XX^T)^t\Omega^{(b)}] \ \ \mbox{for} \ \ b \in \{1,\ldots,B\}.
\end{equation*}
The $k$-th principal basis left singular vector estimate ($k \in \{1,\ldots,d\}$) is assigned a \textit{stability score}:
\begin{equation*}
\mbox{stab}(t,k,B) = \frac{1}{N} \sum_{j_1=1}^{B-1}\sum_{j_2=j_1+1}^{B}\left|\mbox{cor}\left(U_{j_1k}^{(t)}, U_{j_2k}^{(t)}\right)\right|,
\ \ \ \mbox{where} \ N=\frac{B(B-1)}{2}.
\end{equation*}
Here $U_{rk}^{(t)}$ is the estimate of the $k^{th}$ principal eigenvector
of $X^TX$ based on the $r$-th random projection and
$\mbox{cor}\left(U_{j_1k}^{(t)}, U_{j_2k}^{(t)}\right)$ denotes the
Spearman rank-sum correlation between $U_{j_1k}^{(t)}$ and
$U_{j_2k}^{(t)}$.
Eigenvector directions that are not dominated by independent noise are
expected to have higher stability scores. When the data has approximately
low-rank we expect a sharp transition in the eigenvector stability between
the directions corresponding to signal and to noise. In order to estimate this change point,
we apply a non-parametric location shift test (Wilcoxon rank-sum) to each
of the $d_{\max} - 2$ stability score partitions of eigenvectors with larger versus
smaller eigenvalues. The subset of principal eigenvectors that can be
stably estimated from the data for the given value of $t$ is determined
by the change point with smallest p-value among all $d_{\max} - 2$
non-parametric tests.
\begin{equation*}
\hat{d}_{t} = \argmin_{k \in\{2,\ldots,d_{\max} - 1\}} \mbox{p-value}(k,t)
\end{equation*}
where $\mbox{p-value(k,t)}$ is the p-value from the Wilcoxon rank-sum test
applied to the $\{\mbox{stab}(t,i,B)\}_{i=1}^{k-1}$ and
$\{\mbox{stab}(t,i,B)\}_{i=k}^{d_{\max}}$.

\paragraph{Estimation of $t^{*}$:}\label{sec:adapt_t}
In this section we describe a procedure for selecting an optimal value for  
$t \in \{1, \ldots, t_{\max}\}$ based on the reconstruction accuracy under 
Bi-Cross-Validation for SVD \citep{Owen2009}, using the generalized Gabriel 
holdout pattern \citep{Gabriel2002}. The rows and columns of the input matrix
are randomly partitioned into $r$ and $c$ groups respectively, resulting in a total of
$r \times c$ sub-matrices with non-overlapping entries. 
We apply \textit{Adaptive Randomized SVD} to factorize the training data 
from each combination of $(r-1)$ row blocks and $(c - 1)$ column blocks.
In each case the submatrix block with the omitted rows and columns is
approximated using its modified Schur complement in the training data. 
The cross-validation error for each data split corresponds to the Frobenius norm of the
difference between the held-out submatrix and its training-data-based estimate.
For each sub-matrix approximation we estimate the rank
$d^*= d(t)$ using the stability-based approach from the previous section.
As suggested in previous work~\citep{Owen2009}, we fix $c=r=2$, in which case
Bi-Cross-Validation error corresponding to holding out the top left block
$A$ of a given block-partitioned matrix $\mattwo{A}{B}{C}{D}$ becomes
$||A - BD^{+}C||^2_F$. Here $D^{+}=VS^{-1}U^T$ is the Moore-Penrose
pseudoinverse of $D\eqsvd USV^T$.
For fixed value of $t$ we estimate $d(t)$ and factorize $D^{+}$ using
\textit{Adaptive Randomized SVD(t, d(t), $\Delta=10$)} and denote
the Bi-Cross-Validation error by $\text{BiCV}(t, A)$. The same process
is repeated for the other three blocks B, C, and D. The final error and
rank estimates are defined to be the medians across all blocks and are
denoted as $\text{BiCV}(t)$ and $\hat{d}_t$, respectively.
We optimize over the full range of allowable values for $t$ to arrive at the final estimates
\begin{eqnarray*}
\hat{t}_{\text{BiCV}}^* &=& \argmin_{t \in \{1,\ldots,t_{\max} \}} \text{BiCV}(t)\\
\hat{d}_{\text{BiCV}}^* &=& \hat{d}_{\hat{t}_{\text{BiCV}}^*}.
\end{eqnarray*}

\subsection{Fast linear mixed models}
\label{sec:gegin}

Multivariate linear mixed models (LMMs) have been a workhorse in statistical genetics and quantitative genetics because they allow for the regression of explanatory on outcome variables while modeling relatedness between samples \citep{Henderson,Price11,Korte12}. In the context of mapping traits LMMs are used to correct for unobserved sources of statistical confounding in genetic association studies---particularly the presence of population structure amongst samples.

The linear mixed models in this paper take the form  
$$y = x\beta + Z\mu + e,$$
where $y$ is the response, $x$ are the covariates, $\beta$ are the fixed effects,  $Z$ is the random effects design matrix, and
and $\mu$ captures unobserved covariates with $\mbox{var}(\mu) = \sigma_g^2K$ and $\mbox{var}(e) = \sigma_e^2I$. Here, $K$ is the estimated kinship or genetic relatedness matrix, and $\sigma_g^2$ is the proportion of variance in the phenotypes explained by genetic factors. In the standard application to genetic association studies, the response $y$ is the quantitative trait across individuals and $x$ is a $p$-vector of genotypes of those individuals across many genomic locations. The statistical test then is to determine whether the coefficient for a particular genetic variant, $\beta_j$, is equal to zero, indicating no genetic association, or alternatively whether $\beta_j \neq 0$, indicating support for a possible genetic association at this genetic locus with the trait. The likelihood for this model follows from a multivariate normal distribution:
\begin{eqnarray*}
y &\sim& N(X\beta,V)\\
V &=& \sigma_g^2 + \sigma_e^2\\
l(y) &=& -\frac{1}{2}[n\text{log}(2\pi)+\text{log}|V|+(y-X\beta)^T V^{-1} (y-X\beta)].
\end{eqnarray*}
The standard procedure to correct for sample (population) structure using an LMM proceeds in three main steps: (1) empirical estimation (construction) of the genetic relatedness matrix (GRM) amongst individuals based on held-out genetic data; (2) estimation of variance components to account for sample structure contributing to response; and (3) computation of an association statistic at each genotype position. 

Many methods have focused on reducing the complexity of one or more of these steps to scale LMMs to current study sizes. The EMMA method~\citep{kang2008} uses spectral decomposition to jointly estimate the likelihood of the response and random effects together instead of obtaining estimates of random effects \emph{a priori}, which incurs a greater computation cost. EMMAX~\citep{kang2010} improves upon EMMA by estimating variance components only once (independent of the number of genetic variants tested), making the assumption that most variants have a small effect on the phenotype (the pylmm software used here, along with our adaption, is based on EMMAX). FaST-LMM~\citep{lippert2011}, in addition to using a heuristic low-rank spectral decomposition to construct an orthogonal representation of the GRM, uses rotations of the genetic variants and response to perform computation of the likelihood function in linear time. Further methods~\citep{lippert2013} use more systematic methods for subsetting genetic variants that are relevant to the phenotype of interest in order to produce accurate variance component estimations in a LMM.

Our contribution to accelerating parameter estimates in this model is to use ARSVD to address the complexity of estimating the design matrix $Z$. In general, the GRM is computed using a sample-by-sample covariance of the design matrix ($n \times p$), which has a computational complexity of ${\cal O}(np^2)$ to build, where typically $p$ dominates $n$ in genomic studies. Decomposing this matrix into an exact orthogonal representation using an eigendecomposition has complexity ${\cal O}(n^3)$ in the worst case. We forgo explicit computation of the GRM, and instead perform ARSVD on the original design matrix, with a computational complexity of ${\cal O}(n p \times d_{\max} + n \times d^2_{\max} )$. By using ARSVD to decompose the design matrix, we automatically detect low-rank structure present in the GRM, and we avoid the need to manually or heuristically subset the data to achieve a low-rank representation. We note in the simulated data and in the genomic study results that this has both a substantial acceleration in the computational speed and also the implicit regularization of the design matrix $Z$ from our approach improves the type I error substantially.

\section{Results on simulated data}\label{sec:results}
We use real and simulated data to demonstrate three major contributions of this work: 
\begin{enumerate}
\item In the presence of interesting \textit{low-rank} structure in the data, 
the randomized algorithms tend to be much faster than the exact methods 
with minimal loss in approximation accuracy.

\item The \textit{rank} and the \textit{subspace} containing the information 
in the data can be reliably estimated and used to provide efficient
solutions to dimension reduction methods based on the truncated (generalized) 
eigendecomposition formulation.

\item The randomized algorithms implicitly impose regularization, which can be adaptively controlled in a computationally
efficient manner to produce improved out-of-sample performance.
\end{enumerate}

\subsection{Unsupervised dimension reduction}\label{sim:unsup}
We begin with unsupervised dimension reduction of data with low-rank
structure contaminated with Gaussian noise, and we focus on evaluating the
application of \textit{Adaptive Randomized SVD} 
for PCA (see Section \ref{sec:rsvd}). In particular, we demonstrate that the proposed method estimates the
sample singular values with exponentially decreasing relative error in $t$.
Then we show that achieving similar low-rank approximation accuracy to a
state-of-the-art Lanczos method requires the same run time complexity,
which scales linearly in both dimensions of the input matrix. This makes our 
proposed method applicable to large data matrices. Lastly, we demonstrate the ability 
to adaptively estimate the underlying rank of the data, given a coarse upper 
bound. For the evaluation of the randomized
methods, based on all our simulated and real data, we assume a default
value for the oversampling parameter $\Delta=10$.

\paragraph{Simulation setup}
The data matrix $X \in \RR^{n \times p}$, is generated as follows:
$X=USV^T+E$, where $U^TU=V^TV=I_{d^{*}}$. The $d^{*}$ columns
of $U$ and $V$ are drawn uniformly at random from the corresponding
unit sphere and the singular values S = $\mbox{diag}(s_1,\ldots,s_{d^*})$
are randomly generated starting from a baseline value,
which is a fraction of the maximum noise singular value,
with exponential increments separating consecutive entries:
\begin{eqnarray*}\label{eqn1_sim:unsup}
s_j &=& s_{j-1} + \nu_{j}, \ \mbox{for}\ j \in \{2,\ldots,d^{*}\} \\
\nu_j &\iid& \mbox{Exp}(1), \ \ \ \nu_0 = \kappa \times s^{(E)}_1.
\end{eqnarray*}
The noise is iid Gaussian: $E_{ij} \iid N\left(0, \frac{1}{n}\right)$. The sample
variance has the SVD decomposition $E \eqsvd U_E S_EV^T_E$, where
$S_E = \mbox{diag}\left(s^{(E)}_1,\ldots,s^{(E)}_{min(n,p)}\right)$ are the
singular values in decreasing order.
The signal-to-noise relationship is controlled by $\kappa$,
large values of which correspond to increased separation
between the signal and the noise.

\paragraph{Results}
In our first simulation we set $\kappa=1$ and assume the
rank $d^{*}$ to be given and fixed to $50$. The main focus
is on the effect of the regularization parameter $t$ controlling
the singular value shrinkage, with larger
values corresponding to stronger preference for the higher
versus the lower variance directions. The simulation uses an
input matrix of dimension $2,000 \times 5,000$.
Studying the estimates of the 
percent relative error of the \textit{singular values} averaged over 
ten random data sets, we observed exponential convergence 
to the sample estimates with increasing $t$ (Table \ref{tab1}). This suggests that the variability
in the sample data directions is approximated well for
large data sets at the cost of few data matrix multiplications.
\begin{table}[ht]
\begin{center}
\begin{tabular}{|c|ccccc|}
\hline
t & 1 & 2 & 3 & 4 & 5 \\
\hline
\% relative error & $26.1\pm 0.2$ & $8.8\pm 0.2$ & $3.0\pm 0.1$ & $1.0\pm 0.1$ & $0.3\pm 0.0$\\
$\mbox{log}_{10}$[\mbox{mean}] & $4.7$ & $3.1$ & $1.6$ & $-0.1$ & $-1.9$ \\
\hline
\end{tabular}
\caption{{\bf Singular value estimation using Adaptive Randomized SVD.} We report 
the mean relative error ($\pm 1$ standard error) averaged across ten random 
replicates of dimension $2,000 \times 5,000$, with rank 50. Linear increase in 
the regularization parameter $t$ results in exponential decrease in the 
percent relative error.}
\label{tab1}
\end{center}
\end{table}

\begin{figure}[h!tb]
\centering
\includegraphics[scale=0.5]{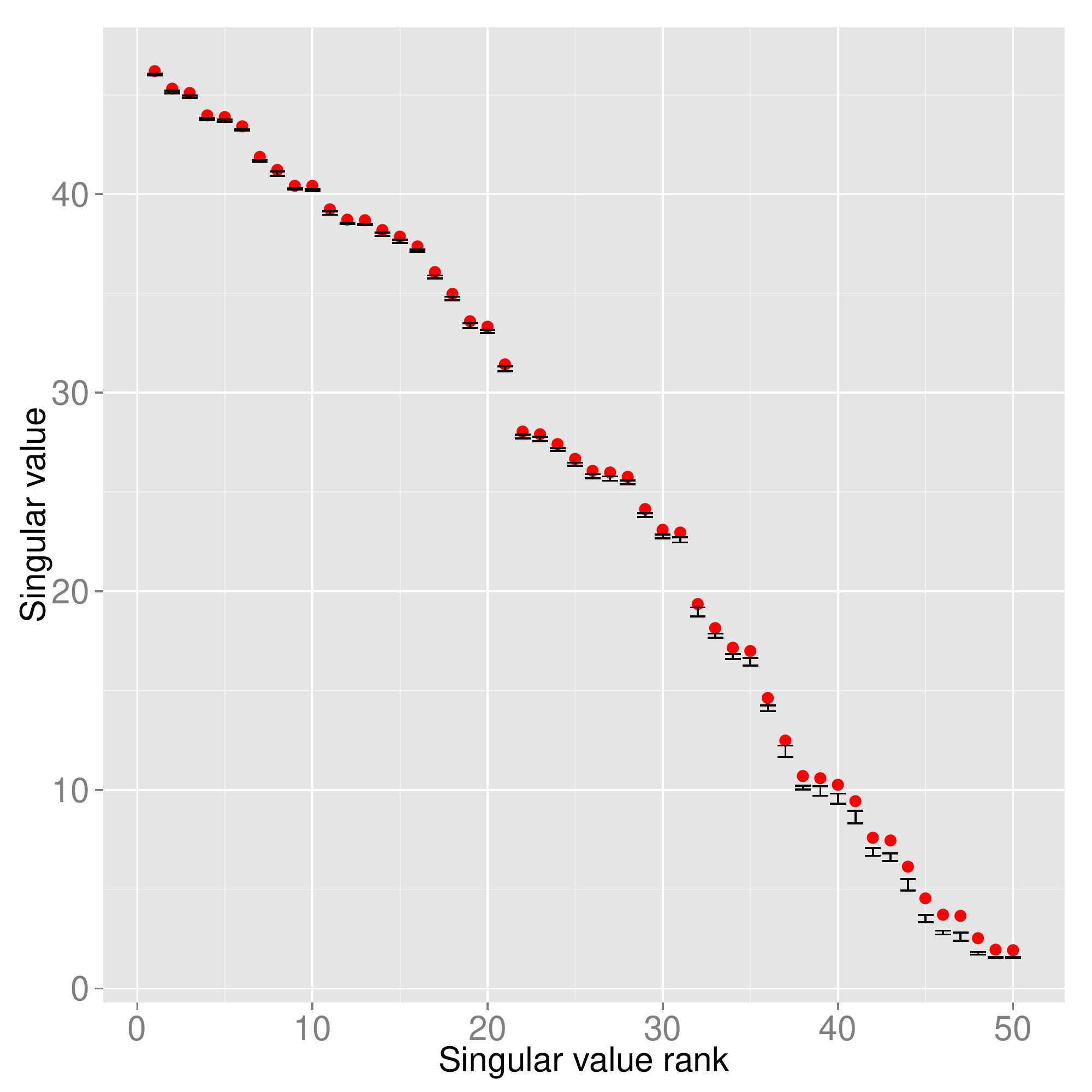}
\caption{{\bf Singular value accuracy of ARSVD.}
Simulation results from random data sets of dimension $n = 1,000$ and $p = 1,000$. Singular values ($\sigma_i$) obtained from
exact SVD are point estimates in red. Singular value confidence intervals are twice the standard error in estimates
from ten estimations using ARVSD.}
\label{fig:sing_accuracy}
\end{figure}

In addition to minor relative error compared with exact approaches, we further investigate the source of putative error
(Figure \ref{fig:sing_accuracy}). Data were generated from $n = 1,000$ and $p = 1,000$ with true rank $d^*=50$. We note
our method is extremely precise for the greatest singular values but has a slightly increased decay rate of singular
values compared to exact methods, where less overall variance is captured in directions present in the data that
are more influenced by noise. Our method has larger standard error and tends to slightly underestimate the smallest singular values,
which we interpret as providing regularization to directions of noise, as further explored in Section \ref{sec:popsims}.

\noindent Similar ideas have been proposed before in the absence of
randomization. Perhaps the most well-established and computationally
efficient methods are based on Lanczos Krylov Subspace
estimation, which also operate on the data matrix only through
matrix multiplies \citep{saad1992numerical, arpack, stewart2001matrix, irlba}.
These methods are also iterative in nature and tend to converge
quickly with run time complexity, typically scaling as the product of the
data dimensions $O(qnp)$, with $q$ small.

In order to further investigate the practical run time behavior of
\textit{Adaptive Randomized SVD} we used one such state-of-the art
low-rank approximation algorithm, Blocked Lanczos \citep{irlba}. Here
we use the same simulation setting as before with fixed $d^{*}=50$, but we vary the
regularization parameter $t$ to achieve comparable percent relative low-rank
Frobenius norm reconstruction error (to 1 d.p.). 
In
Table \ref{tab2} we report the ratio of the run times of the two
approaches based on ten random data sets.
Notice that the relative run time remains approximately constant with simultaneous
increase in both data dimensions, which suggests similar order of complexity for both
methods. In the presence of low-rank structure,
\textit{Adaptive Randomized SVD} is feasible for larger data problems
where full-factorization methods (e.g. \citep{Anderson:1990:LAPACK}) are not; moreover, \textit{Adaptive Randomized SVD}
achieves good approximation accuracy for the top sample singular
values and singular vectors.
\begin{table}[ht]
\begin{center}
\begin{tabular}{|r|ccccc|}
\hline
n + p & 6,000 & 7,500 & 9,000 & 10,500 & 12,000\\
\hline
relative time & $2.5 \pm 0.05$& $1.84 \pm 0.03$& $1.82 \pm 0.03$& $1.83 \pm 0.02$ & $1.84 \pm 0.04$\\
\hline
\end{tabular}
\caption{{\bf Run time ratio between \textit{Adaptive Randomized SVD} and Lanczos.} Results are across
different data dimensions $n$ and $p$ and rank $d^{*}=50$. The Lanczos implementation is
from the {\tt irlba} CRAN package~\citep{irlba}. The \textit{Adaptive Randomized SVD}
was run until the percent relative low-rank reconstruction error was equal to the Lanczos error
to 1 d.p. 
For each consecutive simulation scenario $n$ is incremented by $500$ and $p$ is
 incremented by $1,000$. We report the sample mean and standard error based on ten random 
replicate data sets.}
\label{tab2}
\end{center}
\end{table}

\begin{figure}[h!tb]
\begin{minipage}{.4\textwidth}
\centering
\includegraphics[scale=0.5]{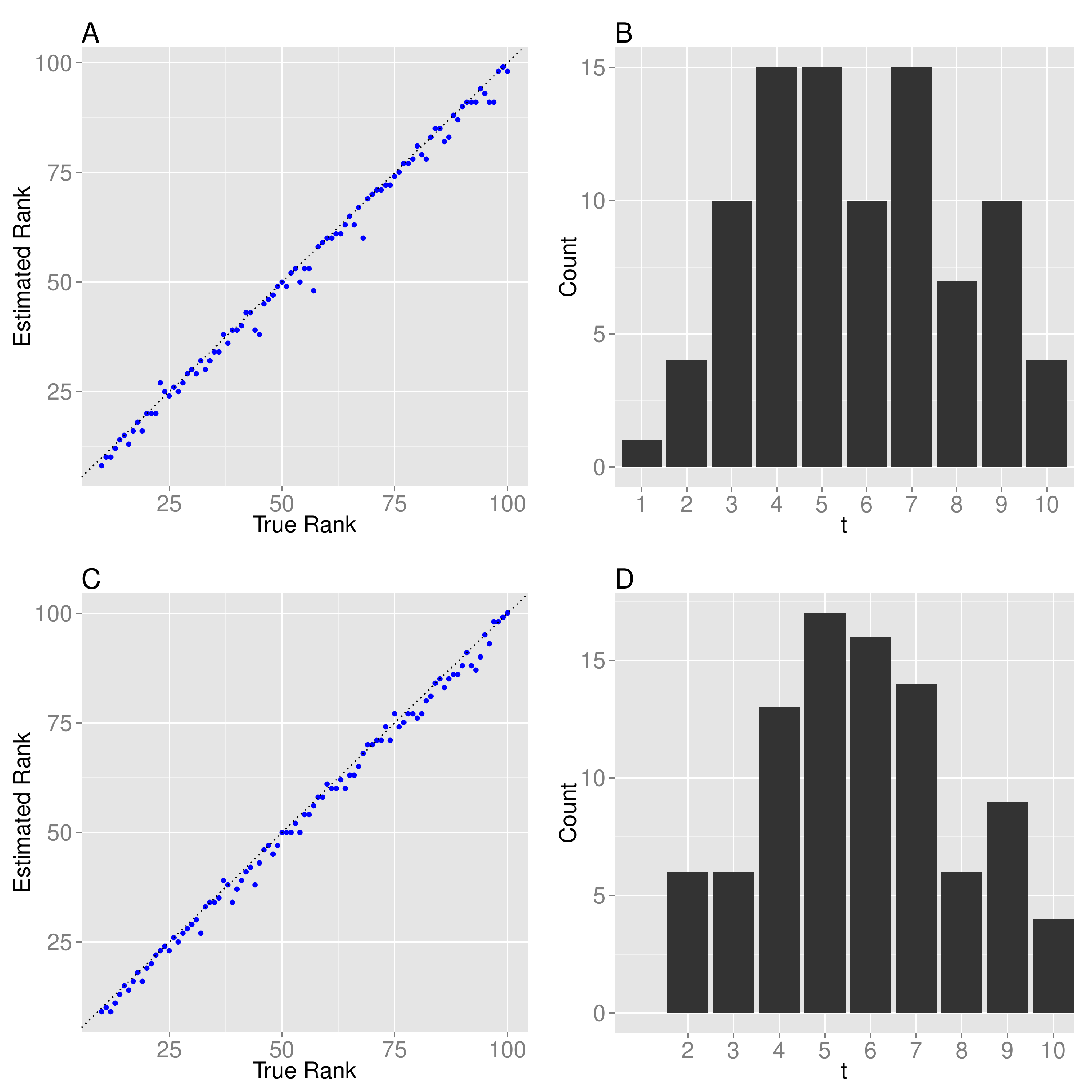}
\end{minipage}
\caption{{\bf Rank estimation of ARSVD.}
Simulation results from 50 random data sets of dimension $n = 10,000$ and $p = 10,000$,
with true rank $d^{*} \iid \mbox{Uniform}[10,50]$. Panel A,C: True rank ($d^{*}$) on the x-axis, and estimated rank ($\hat{d}^{*}$) on the y-axis for signal-to-noise-parameters $\kappa=1$, $\kappa=2$, respectively. Panel B,D: estimations of $t^*$ with max $t$ set to ten for signal-to-noise parameters $\kappa=1$, $\kappa=2$, respectively.}
\label{fig:rank_benchmark}
\end{figure}

In many modern applications the data have latent
low-rank structure of unknown dimension. In Section \ref{rankest_unsup}, we address the
issue of estimating the rank using a stability-based approach. Next, we studied the
ability of our randomized method to perform rank estimation to identify the
dimensionality in simulated low-rank data. For that purpose, we generated $50$ random
data sets, $n=1,000, p=1,000$, $d^{*} \iid \mbox{Uniform}[10,50]$,
and $\kappa=1, 2$. We set an initial rank upper bound estimate to be
$d^{*} * 2$ and used the adaptive
method ARSVD (Section \ref{rankest_unsup}) to estimate both optimal $t^{*}$ and 
the corresponding $d^{*}$. Figure \ref{fig:rank_benchmark} 
plots the \textit{estimated} versus the \textit{true rank} and the corresponding estimates of the
regularization parameter $t^{*}$ for two different signal-to-noise scenarios. In both scenarios, the 
rank estimates showed good agreement with the true rank values. When $\kappa=1$
the smallest ``signal" direction has the same variance as the largest variance ``noise" 
direction, which causes our approach to slightly underestimate the largest ranks. 
This is due to the fact that the few smallest variance signal directions tend to be 
difficult to distinguish from the random noise and hence less stable under random 
projections. We observe that our approach tends to select small values for $t^{*}$, 
especially when there is a clear separation between the signal and the noise
 (right panel of Figure \ref{fig:rank_benchmark}).
This results in a reduced number of matrix multiplications and hence fast computation.

\begin{figure}[h!tb]
\centering
\includegraphics[scale=0.43]{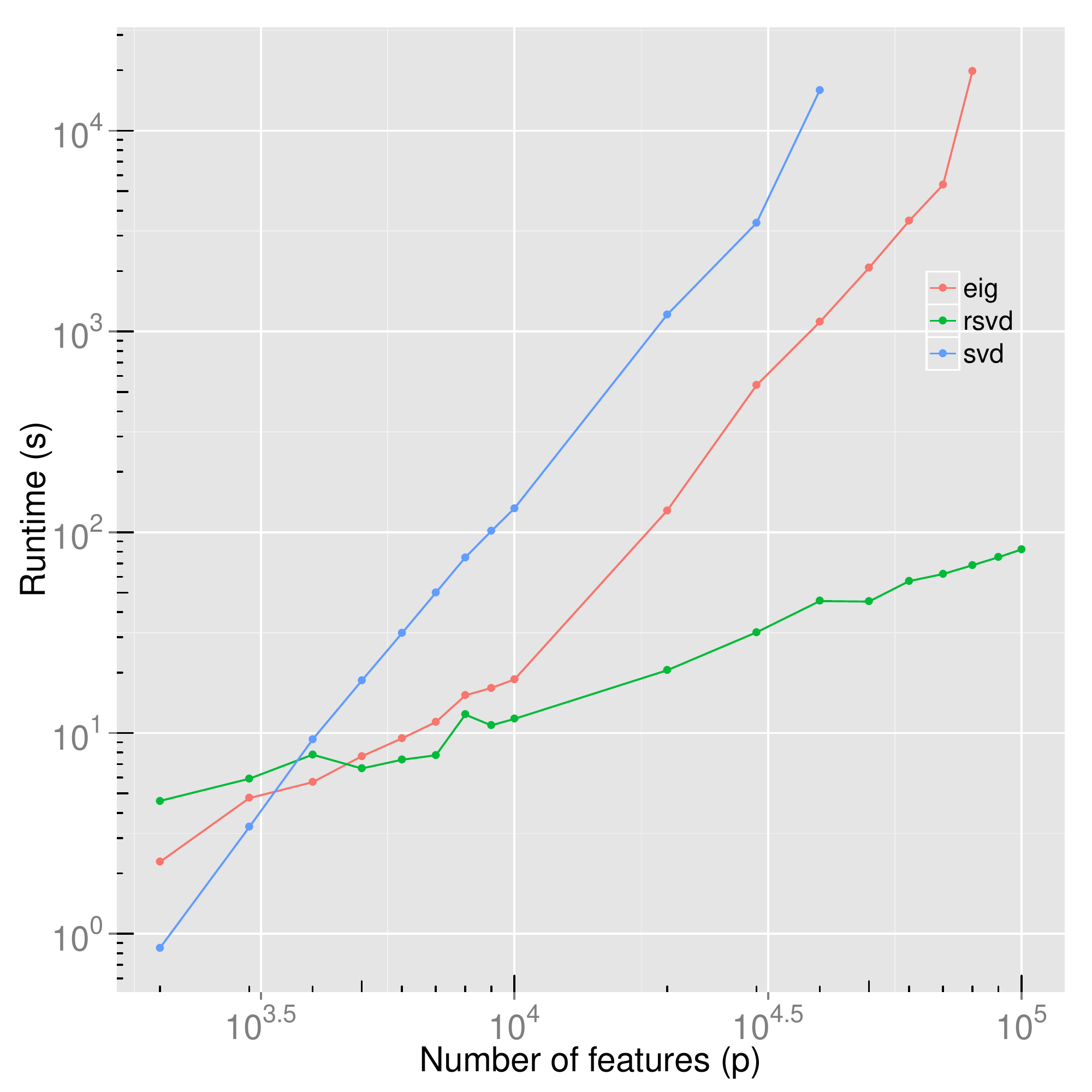}
\caption{{\bf CPU time for three matrix decomposition methods with respect to number of features.}
Simulation results from pseudo-random data sets with range of number of features, $p =$ 2,000 to 100,000 (rsvd), 80,000 (eig), and 40,000 (svd). The number of samples, $n$ is one tenth the number of features. 
Run time is reported in terms of CPU-time (seconds) versus total number of features, $p$. Both axes scale with the log of the axis metric.
SVD is performed on the complete $n \times p$ matrix.
Eigendecomposition is performed with the sample covariance, that is, after multiplying design matrix $X$ by $\text{X}^\intercal$ and performing a decomposition on the resulting $n \times n$ matrix.
RSVD estimates the top 100 principal components.}
\label{fig:comp_results}
\end{figure}


\subsection{Results on simulated data with latent population structure}\label{sec:popsims}

To simulate a genomic association study with latent population structure, we generate genotypes and phenotypes that are conditionally independent given the true population structure, as described in~\citep{mimno2014}. Each individual's genotype is admixed, containing a proportion of ancestry from each population ${1,...,K}$. The admixture proportions for each individual $i$, are drawn $\theta_i \sim Dir_K(\alpha)$. Since we only consider two alleles at each position in the genome, the frequency of each variant $j = {1,...,P}$ in population $k$ is drawn $\phi_{j,k} \sim Beta(1,1)$ representing a uniform allele frequency. To generate latent assignments of each variant to a population, we create variant-population assignments by generating $z_{1,i,j}$ and $z_{2,i,j}$ from $Mult(\theta_i)$. The final generation of genotypes draws each variant $j$ from individual $i$ as $Bin(\phi_j,k)$. To generate a phenotype (response) vector for the $k^{th}$ population, we draw $y_i \sim Bern(0.5\theta_{k,i} + 0.1(1-\theta_{k,i}))$.

Results are generated for four distinct scenarios representing data with differing structure. The simulations presented here have no true associations between genotypes (features) and phenotypes (response), but are dependent conditional on latent population structure. This situation mimics the commonly observed confounding effect in genomic data, and motivates the need for LMMs and regularization in genome-wide studies. Given the simulations are entirely under the null hypothesis of no association between genotype and phenotype, it is expected that p-values follow a uniform distribution; we consider any result exceeding some $\alpha$-threshold a false positive, occurring at rate $\alpha$.

\begin{figure}[h!tb]
\begin{minipage}{.4\textwidth}
\centering
\includegraphics[scale=0.5]{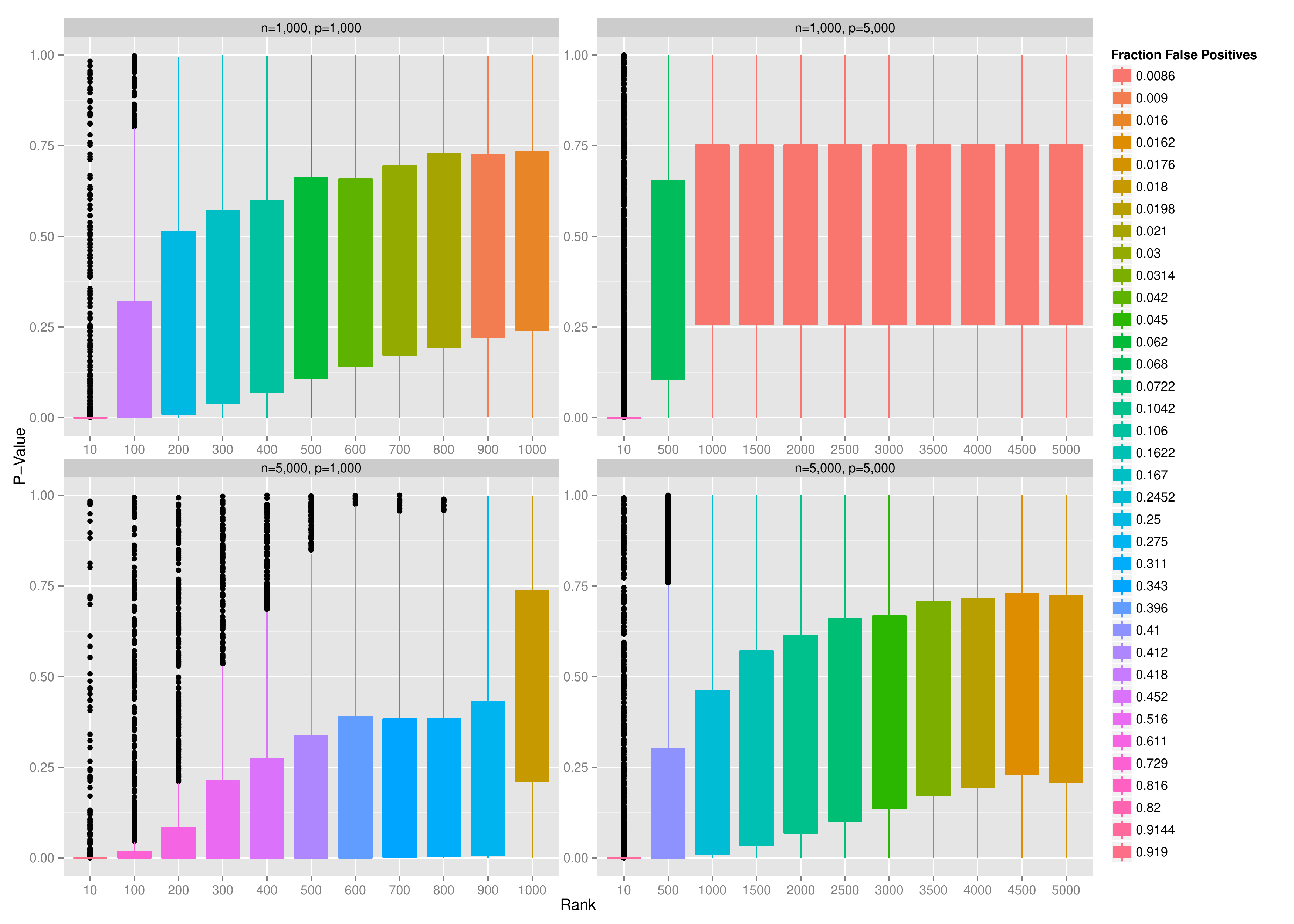}
\end{minipage}
\caption{{\bf Regularization simulations.}
X-axis is latent rank used to project the original matrix into a lower subspace. Y-axis is p-values across $p$ tests for association. Each box plot is colored by the fraction of false positives in the association results for a given rank. Top left: $n=100$ and $p=100$. Top right: $n=100$ and $p=1,000$. Bottom left: $n=1,000$ and $p=100$. Bottom right: $n=1,000$ and $p=1,000$.}
\label{fig:regularization}
\end{figure}

In each simulation, the p-value distribution from an LMM association without ARSVD is reported as the full data matrix rank (the largest rank on the x-axis).
The top left simulation uses $n=1,000$ (samples or genetic variants) and $p=1,000$ (responses or values of the phenotype), representing a situation with few samples. Our method performs some slight regularization beyond that of a standard LMM with rank close to full data rank (Figure~\ref{fig:regularization}).
The next simulation (top right, $n=1,000$ and $p=5,000$) is the most common case in genomics with $p \gg n$, or a large set of features compared to responses. Our method performs equally well as current methods but at a fraction of the computational cost, as illustrated by the lowest-rank estimation that still yields uniform p-values and low false positive rate.
In the bottom left panel we generate a case of an over parametrized model ($n=5,000$ and $p=1,000$), for which our method produces significant false positives compared to a standard LMM; however, we note that if we know \emph{a priori} that our model is over parametrized, we can modify our decomposition to perform in the sample space instead of feature space and achieve comparable results to a standard LMM.
In the last setting (bottom right, $n=5,000$ and $p=5,000$) again our method performs similarly to the standard method but with gains in computational efficiency. In general, we expect the tradeoff between computational efficiency and accuracy to be dependent on the data analyzed. In the case of structured genomic data, we report in these simulations and in Section~\ref{real:data:model} that massive computational savings are accompanied by numerical accuracy.

\section{Applications to large scale genomic data}\label{real:data:model}

\subsection{Results on WTCCC association mapping}

We test our approach that incorporates ARSVD in LMM association studies on the Wellcome Trust Case Control Consortium (WTCCC) data~\citep{burton2007} that includes 4,684 individuals (half case individuals with Crohn's disease and half control individuals without Crohn's disease) and 478,765 genetic variants across the 22 autosomal chromosomes in the genome. We compare our method ARSVD incorporated in the next generation LMM software after EMMAX (pylmm) to GEMMA. Performing ARSVD on the whole genome took 82.2 secs, while the traditional eigendecomposition of the covariance matrix in pylmm took 88 mins 23.9 sec. In order to most accurately control for the test statistic computed in a LMM and achieve maximal statistical power, it is suggested that a covariance matrix is constructed once per (22) chromosomes, performed by holding out the test chromosome and concatenating the remaining chromosomes~\citep{yang2014}. Our method performs the 22 decompositions in a total of 5 mins 4.8 secs, while the traditional method performs the same decomposition in 4 hrs 24 mins.

GEMMA is executed with the {\tt -lmm} option to most closely resemble the analysis that pylmm performs. Despite having discrete phenotypes (disease status \{0,1\}), we note that estimating coefficients $\beta$ (effect sizes) with linear regression and logistic regression have nearly indistinguishable impacts on computing association statistics~\citep{zhou2013}. Estimates of $\beta$ for pylmm and GEMMA have a high correlation, yet GEMMA shows no enrichment for significant p-values. On the other hand, the distribution of p-values estimated from pylmm is closer to what we expect from an analysis on true associations, showing that our method, while providing some regularization, does not impact enrichment of true signal.

\begin{figure}[h!tb]
\centering
\includegraphics[scale=0.38]{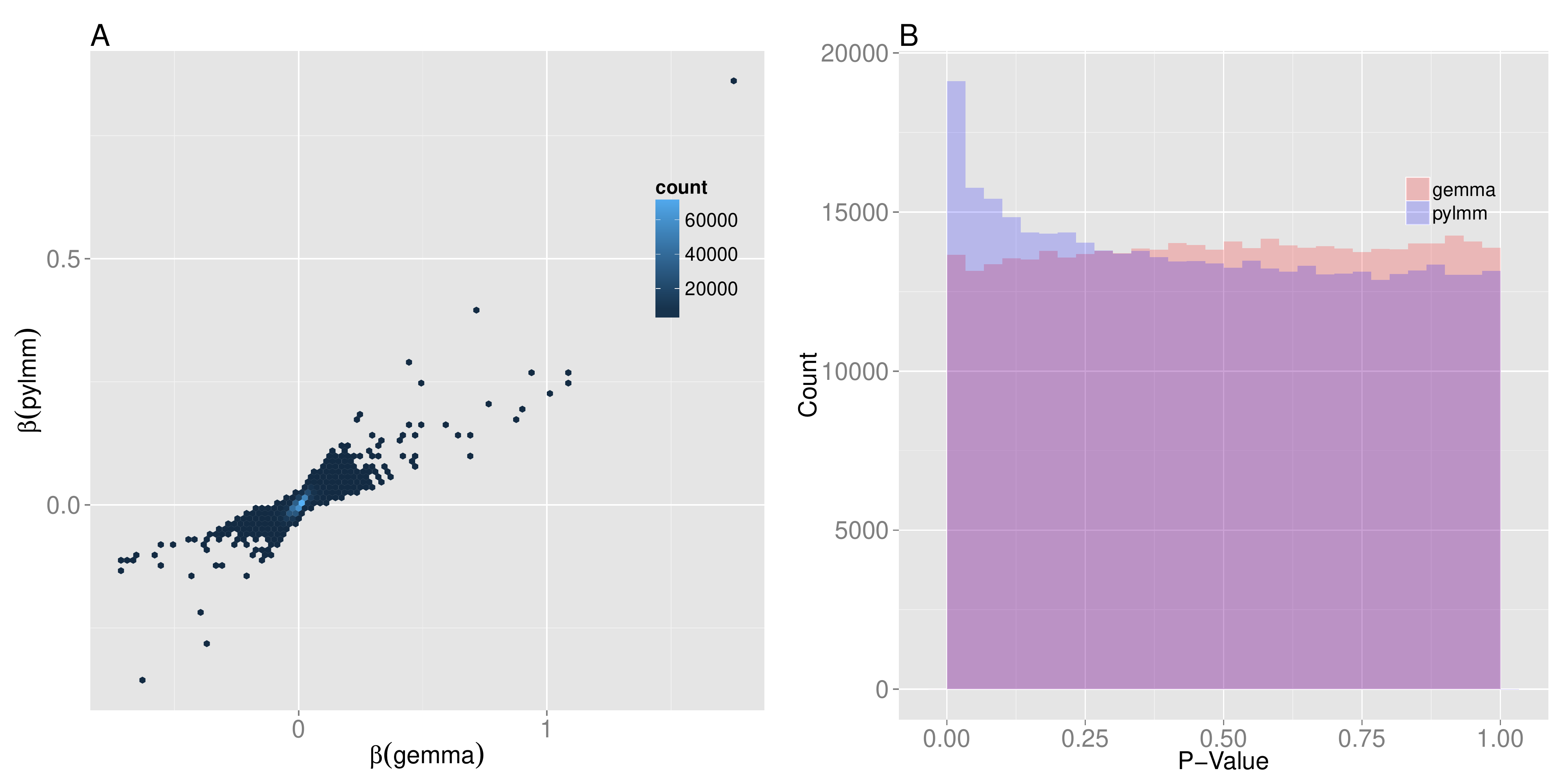}
\caption{{\bf Genome-wide association study results for pylmm and GEMMA.}
Panel A: $\beta$ of every analyzed variant in the genome, GEMMA versus pylmm. Panel B: Histogram of p-values for each method.}
\label{fig:gwas_results}
\end{figure}

We investigated the most significant genetic associations identified by our method and compared to previous results from a large-scale meta-analysis of Crohn's disease. This meta-analysis had a much larger sample size than the WTCCC: it included 6,333 affected individuals (cases) and 15,056 controls and followed up the top association signals in 15,694 cases, 14,026 controls, and 414 parent-offspring trios. Table~\ref{table:gwas_hits} reports the curated list of associated genetic variants reported in this meta-analysis through various sources~\citep{listgarten2012}. This includes genetic variants identified by meta-analysis~\citep{franke2010}, the original WTCCC study~\citep{burton2007}, or those that were confirmed more recently via several methods and known to correspond to a region associated with Crohn's disease risk~\citep{listgarten2012}. For each analysis type, we report the number of variants found significant using this analysis, the subset of those variants that lie within the top 0.5\% of our results, and the subset of variants that reach significance at a local false discovery rate (lfdr) of 5\%~\citep{strimmer2008}.

Furthermore, we identify several genetic variants associated with Crohn's that were previously unidentified. In particular, there are a few genetic variants within the 3.6MB region that defines the MHC on chromosome 6 in the human genome~\citep{the1999} at an lfdr threshold of 10\%.
In particular, genetic variant rs9269186 ($p\leq 7.03\times 10^{-7}$) passes below an lfdr threshold of 1\%, and lies within 5KB of the start of the \emph{HLA-DRB5} gene, putatively acting to regulate transcription of this protein-coding gene that produces a membrane-bound class II molecule. The \emph{HLA-DRB5} protein is an antigen that has an important role in the human immune system, and thus may play a role in an autoimmune disorder such as Crohn's disease.

\begin{table}[ht]
\begin{center}
\begin{tabular}{|l|c|c|c|}
\hline
source & significant variants & top 0.5\% & lfdr 5\% \\
\hline
All & 151 & 61 & 35\\
\hline
MA + WTCCC & 93 & 48 & 30\\
\hline
MA & 81 & 47 & 29\\
\hline
\end{tabular}
\caption{{\bf Enrichment of genetic associations found in our results among related Crohn's disease studies.} Number of genetic variants associated with Crohn's disease risk identified by previous studies that are present in the top 0.5\% of the results obtained from our method. MA = meta-analysis. MHC = major histocompatibility complex, which is a region previously identified as associated with Crohn's and autoimmune diseases more generally. WTCCC = original analysis performed on data. All = MT + MHC + WTCCC.}
\label{table:gwas_hits}
\end{center}
\end{table}

\newpage

\section{Discussion}
Massive high dimensional data sets are ubiquitous in modern data analysis settings.
In this paper we address the problem of tractably and accurately performing
eigendecompositions when analyzing such data. The
main computational tool we use is based on recent randomized algorithms
developed by the numerical analysis community. Taking into account the presence of
noise in the data we provide an adaptive method for estimation of both the rank $d^*$
and number of Krylov iterations $t^*$ for the randomized approximate version of SVD.
Using this adaptive estimator of low-rank structure, we implement efficient algorithms for
PCA and linear mixed models.
Perhaps, the most interesting statistical observation is related to the
regularization that randomization implicitly imposes on the resulting eigendecomposition estimates.

In simulated experiments we show high accuracy in recovering the true (latent) rank of 
matrices with low-rank substructure, without the need for many Krylov iterations. Additionally, 
we show in simulations that our method performs implicit regularization 
and improves quantitative properties of results under various settings of data structure.
Furthermore, our results on modern genome-wide association studies show that our approach to using ARSVD in linear mixed models fills a critical need for 
methods with computational efficiency that do not sacrifice the desirable statistical properties of 
these tests, but instead have the potential to improve upon these properties.
Some important open questions still remain:
\begin{enumerate}
\item[(1)] There is need for a theoretical framework to quantify what generalization
guarantees the randomization algorithm can provide on out-of-sample
data, and the dependence of this bound on the noise and the structure in
the data on one hand and on the parameter settings on the other.
\item[(2)] A probabilistic interpretation of the algorithm could contribute additional insights
into the practical utility of the proposed approach under different assumptions. In particular
it would be interesting to relate our work to a Bayesian model with posterior modes that
correspond to the subspaces estimated by the randomized approach.
\item[(3)] The implicit regularization on latent factors imposed by ARSVD should be further explored
with respect to the structure of the noise, and its impact on estimates of random effects in LMMs~\citep{runcie2013}.
\end{enumerate}

\acks{SM would like to acknowledge Lek-Heng Lim, Michael Mahoney, Qiang Wu, and Ankan Saha.
SM is pleased to acknowledge support from grants NIH (Systems
Biology): 5P50-GM081883, AFOSR: FA9550-10-1-0436, NSF CCF-1049290,
and NSF-DMS-1209155. BEE is pleased to acknowledge support from grants NIH R00 HG006265 and NIH R01 MH101822.
SG would like to acknowledge Uwe Ohler, Jonathan Pritchard and Ankan Saha.}

\appendix
\section{}

\subsection{Generalized eigendecomposition and dimension reduction}\label{sec:gegein}
The appendix states a variety of dimension reduction methods supervised, unsupervised, and nonlinear can use the ARSVD engine to scale to massive data. The key requirement is a
a formulation of the truncated generalized eigendecomposition problem that can be implemented
by the \textit{Adaptive Randomized SVD} from Section \ref{sec:rsvd}. The dimension reduction methods we will focus on are sliced inverse regression (SIR) and localized sliced inverse regression (LSIR).  

\subsubsection{Problem Formulation.} Assume we are given 
$\Sigma \in \SS^{p}_{++}, \Gamma \in \SS^{p}_{+}$ that
characterize pairwise relationships in the data and let $r \ll \min(n, p)$
be the "intrinsic dimensionality" of the information contained in the data.
In the case of supervised dimension reduction methods this corresponds to
the dimensionality of the linear subspace to which the joint distribution of
$(X,Y)$ assigns non-zero probability mass. Our objective is to find a basis
for that subspace. For SIR and LSIR this corresponds to the span of the
generalized eigenvectors $\{g_1,\ldots,g_{r}\}$ with largest eigenvalues 
$\{\lambda_{\max} = \lambda_1 \le \ldots \le\lambda_{r}\}$:
\begin{eqnarray}\label{eqn:gsvd:exact1}
\Gamma g=\lambda \Sigma g.
\end{eqnarray}
An important structural constraint we impose on $\Gamma$, which is
applicable to a variety of high-dimensional data settings, is that it has 
low-rank: $r \le d^{*} \equiv \mbox{rank}(\Gamma) \ll p$. 
It is this constraint that we will take advantage of in the 
randomized methods. In the case of $\Sigma = \mbox{I}$ (unsupervised case)
$r=d^{*}$.

\subsubsection{Sufficient dimension reduction}\label{sec:sdr}
Dimension reduction is often a first step in the statistical analysis of
high-dimensional data and could be followed by data visualization or
predictive modeling. If the ultimate goal is the latter, then the statistical
quantity of interest is a low dimensional summary $Z \equiv R(X)$ which
captures all the predictive information in $X$ relevant to $Y$:
$$ Y = f(X) + \varepsilon = h(Z) + \varepsilon, \quad X\in \RR^{p}, Z \in \RR^{r}, \ r \ll p.$$
Sufficient dimension reduction (SDR) is one popular approach for estimating $Z$, which
\citep{Li:1991, Cook:Weis:1991, Li:1992, Li2005,Nilsson07, Sugiyama2007, Cook2007, lsir:2010}.
In this appendix we focus on linear SDRs:
$G=(g_{1},\ldots,g_{r}) \in \RR^{p \times r} \Rightarrow R(X) = G^T X$, which
provide a prediction-optimal reduction of $X$
$$(Y \mid X) \stackrel{d}{=} (Y \mid G^T X), \quad \stackrel{d}{=}
\mbox{ is equivalence in distribution}.$$

We will consider two specific dimension reduction methods:
Sliced Inverse Regression (SIR) \citep{Li:1991} and Localized Sliced
Inverse Regression (LSIR) \citep{lsir:2010}. SIR is effective
when the predictive structure in the data is global, i.e., there is
single predictive subspace over the support of the marginal distribution of
$X$. In the case of local or manifold predictive structure in
the data, LSIR can be used to compute a projection matrix $G$ that contains this
non-linear (manifold) structure.

\subsubsection{Efficient solutions and approximate SVD}\label{sec:exact:sdr}
SIR and LSIR reduce to solving a truncated generalized eigendecomposition
problem in (\ref{eqn:gsvd:exact1}). Since we consider estimating the dimension reduction
based on sample data we focus on the sample estimators $\widehat{\Sigma}=\frac{1}{n}X^TX$
and $\widehat{\Gamma}_{XY}=X^TK_{XY}X$, where $K_{XY}$ is symmetric and encodes the
method-specific grouping of the samples based on the response $Y$.
In the classic statistical setting, when $n > p$, both
$\widehat{\Sigma}$ and $\widehat{\Gamma}_{XY}$ are positive definite almost surely.
Then, a typical solution proceeds by first sphering the data: $Z=\widehat{\Sigma}^{-\frac{1}{2}}X$,
e.g., using a Cholesky or SVD representation
$\widehat{\Sigma} = \widehat{\Sigma}^{\frac{1}{2}} (\widehat{\Sigma}^{\frac{1}{2}})^T$.
This is followed by eigendecomposition of $\widehat{\Gamma}_{ZY}$ \cite{Li:1991, lsir:2010}
and back-transformation of the top eigenvectors directions to the canonical basis.
The computational time is $O(np^2)$.
When $n<p$, $\hat{\Sigma}$ and $\hat{\Gamma}$ are rank-deficient and a
unique solution to the problem (\ref{eqn:gsvd:exact1}) does not exist. One widely-used
approach, which allows us to make progress in this problematic setting, is to restrict our attention 
to the directions in the data with positive variance. Then we can proceed as before, using 
an orthogonal projections onto the span of the data. The total computation time in this case is $O(n^2p)$.
In many modern data analysis applications both $n$ and $p$ are very large, and
hence algorithmic complexity of $O[\mbox{max}(n,p) \times \mbox{min}(n,p)^2]$ could
be prohibitive, rendering the above approaches unusable.
We propose an approximate solution that explicitly recovers
the low-rank structure in $\Gamma$ using \textit{Adaptive Randomized SVD} from
Section \ref{sec:rsvd}. 
In particular, assume rank $(\Gamma)=d^{*} \ge r$ (where $r$ is the dimensionality
of the optimal dimension reduction subspace).
Then $\Gamma \eqsvd US^2U^T$, where $U \in \RR^{p \times d^{*}}$.
The generalized eigendecomposition problem (\ref{eqn:gsvd:exact1}) solution
becomes restricted to the subspace spanned by the columns of $\Gamma$:
\begin{eqnarray}\label{eqn:symm:gsvd}
S^{-1}U^T \Sigma US^{-1} e &=& \frac{1}{\lambda} e, \quad \ \ \ e \equiv SU^Tg.
\end{eqnarray}
The dimension reduction subspace is contained in the $\mbox{span}(G)$, where
$G = (US^{-1} e_1,\ldots, US^{-1} e_{r})$.

\vskip 0.2in
\bibliography{arsvd}
\end{document}